\documentclass{article}

\usepackage{microtype}
\usepackage{graphicx}
\usepackage{subfigure}
\usepackage{booktabs} 

\usepackage{listings}

\usepackage{hyperref}

\usepackage[accepted]{icml2024}

\usepackage{amsmath}
\usepackage{amssymb}
\usepackage{mathtools}
\usepackage{amsthm}

\usepackage[capitalize,noabbrev]{cleveref}

\theoremstyle{plain}

\theoremstyle{definition}

\theoremstyle{remark}

\usepackage[textsize=tiny]{todonotes}

\icmltitlerunning{Comgra: A Tool for Analyzing and Debugging Neural Networks}

\begin{document}

\twocolumn[
\icmltitle{Comgra: A Tool for Analyzing and Debugging Neural Networks}



\icmlsetsymbol{equal}{*}

\begin{icmlauthorlist}
\icmlauthor{Florian Dietz}{equal,uds}
\icmlauthor{Sophie Fellenz}{rptu}
\icmlauthor{Dietrich Klakow}{uds}
\icmlauthor{Marius Kloft}{rptu}
\end{icmlauthorlist}

\icmlaffiliation{uds}{Spoken Language Systems (LSV), Saarland University, Saarbrücken, Germany}
\icmlaffiliation{rptu}{Machine Learning Group, RPTU, Kaiserslautern, Germany}

\icmlcorrespondingauthor{Florian Dietz}{fdietz@lsv.uni-saarland.de}
\icmlcorrespondingauthor{Sophie Fellenz}{fellenz@cs.uni-kl.de}
\icmlcorrespondingauthor{Marius Kloft}{kloft@cs.uni-kl.de}
\icmlcorrespondingauthor{Dietrich Klakow}{dietrich.klakow@lsv.uni-saarland.de}

\icmlkeywords{Machine Learning, ICML}

\vskip 0.3in
]



\printAffiliationsAndNotice{\icmlEqualContribution} 

\begin{abstract}

Neural Networks are notoriously difficult to inspect.
We introduce comgra, an open source python library for use with PyTorch.
Comgra extracts data about the internal activations of a model and organizes it in a GUI (graphical user interface).
It can show both summary statistics and individual data points, compare early and late stages of training, focus on individual samples of interest, and visualize the flow of the gradient through the network.
This makes it possible to inspect the model's behavior from many different angles and save time by rapidly testing different hypotheses without having to rerun it.
Comgra has applications for debugging, neural architecture design, and mechanistic interpretability.
We publish our library through Python Package Index (PyPI) and provide code, documentation, and tutorials at \url{github.com/FlorianDietz/comgra}.
\end{abstract}

\section{Introduction}

\textbf{Growing Complexity of Neural Networks}
Led by the success of Large Language Models (LLMs), neural networks have massively increased in size in recent years \citep{Naveed2023ACO}.
Moreover, they are not only larger, but also more complex and intricate. Instead of just adding more fully connected layers, modern State of the Art (SOTA) models are often based on a combination of many different types of layers, such as Attention Mechanisms, Convolutional Layers, Normalization Layers, and Residual Connections.

\textbf{Difficulties in Debugging and Optimization}
The complex interactions between their many components make it very difficult to understand how the trained model works internally.
Flagship LLMs frequently differ in small aspects, such as the positioning of the Layer Norm or the decision to use a Dropout Layer.
Optimization is often a long and tedious, iterative process. Design decisions are guided by intuition, which is error prone as often becomes evident when trying to replicate suggested techniques in other architectures.
As a result, debugging and improving neural networks is a very time-consuming and error-prone process, a problem intensifying as model complexity increases.

\textbf{Mechanistic Interpretability}.
The field of Mechanistic Interpretability has emerged as a more ambitious form of traditional interpretability research: It aims to reverse-engineer human-interpretable algorithms from learned network weights \citep{nanda_2022}.
While it suffices to look at correlations between parameters and the model's loss for many of the more traditional interpretability techniques, mechanistic interpretability requires an in-depth look at individual model parameters as well as network activations \citep{Zhang2020ASO,anonymous2024mechanistic}.
Identifying which of the billions of network parameters and intermediate network activations have which effect, for any given sample, poses a significant challenge.




\textbf{Comgra: Computation Graph Analysis}.
In this paper we present Comgra, short for Computation Graph Analysis, a library for debugging and analyzing neural networks in PyTorch \citep{Ansel_PyTorch_2_Faster_2024}.
Comgra can help you to track down anomalies in neural networks, analyze dependencies in them, and rapidly test hypotheses by speeding up your inspections through a convenient Graphical User Interface.
We present a list of existing tools for neural network analysis and summarize their shortcomings. We then introduce comgra to plug this gap in our software toolkit.
We show on a series of usecases how Comgra can be used to help researchers with their work in practice, whether in debugging, in neural architecture design, or in mechanistic interpretability.

\section{Existing Tools for Neural Network Analysis}
\label{section:existing_tools}


\textbf{Tracking Metrics}.
Keeping track of the key metrics of your network is the most basic and fundamental type of analysis. We want to be able to track losses and accuracy values, as well as the distributions and summary statistics of network weights.
\textbf{Tensorboard}\footnote{\label{tensorboard}Tensorboard: \url{github.com/tensorflow/tensorboard}}, which was originally developed for TensorFlow \citep{tensorflow2015-whitepaper}, allows us to track the development of all of these values as training proceeds and enables easy comparisons between different runs. Due to its popularity, it also offers a wide variety of plugins for additional features.


\textbf{Task-specific Visualizations}.
For some types of tasks it is possible to visualize the performance of the model on specific inputs in a human-interpretable way.
\textbf{PyTorch-GradCAM} \citep{jacobgilpytorchcam} makes the performance of neural networks in vision tasks more explainable by visually highlighting how the pixels of an image affect the model's decision.
\textbf{PyTorch-Visdom}\footnote{Visdom: \url{github.com/fossasia/visdom}} is a more generic library that allows you to create custom dashboards with a variety of visualization for different types of data.

\textbf{Attribution and Interpretability}.
It is normally difficult to tell how input features as well as intermediate neurons affect the output of a model.
\textbf{PyTorch-Captum} \citep{kokhlikyan2020captum} provides a library of attribution algorithms to automate this, which makes it easier to interpret your model's behavior.

\textbf{Visualizing the Computation Graph}.
As the size and complexity of a model increases, it can become difficult to keep track of its computation graph, which is necessary to trace the flow of information through the network.
\textbf{Tensorboard}\footref{tensorboard} has a basic feature for this built in, while \textbf{Netron}\footnote{Netron: \url{github.com/lutzroeder/netron}} and Torchlens \citep{torchlens} provide a more detailed visualization.
\textbf{Penzai}\footnote{Penzai: \url{github.com/google-deepmind/penzai}} is a toolkit for visualizing models.

\textbf{Mechanistic Interpretability}.
\textbf{TransformerLens} \citep{nanda2022transformerlens} lets you load a collection of popular language models and makes it easy to inspect and modify their internal activations.
\textbf{Pyvene} \citep{wu2024pyvene} and \textbf{Nnsight} \citep{nnsight} make it easier to inspect and intervene on models.
\textbf{Inseq} \citep{Sarti_Inseq_An_Interpretability_2023} is an interpretability toolkit for sequence generation models.

\section{Comgra}


\textbf{What Parts of the Network Matter?}
When analyzing neural networks, one can look at the data in many different ways: Do we inspect individual tensors, or summary statistics? Raw values, or normalized ones? Do we record at each step during training, or only at specific, important times? Do we inspect random inputs, or specific inputs that are particularly important to us?
Tools that visualize metrics in graphs usually focus on summary statistics and neglect special cases. Conversely, task-specific visualizations usually only focus on a single input image at a time and neglect how these correlate with other inputs.

\textbf{The Need for Flexibility}.
The combination of many intermediate network activations with the many different ways to record them leads to a combinatorial explosion of possibilities. There are simply too many ways to look at the network to record and inspect all of it in a reasonable amount of time.
However, since neural network training is not reversible we need to know in advance what we want to log. If we later realize that we forgot to record some number that we care about, we need to rerun the whole model.
If we care about training dynamics and forgot to record something, we may even have to repeat the entire training process, which may be too computationally expensive to be practical.
At the same time, if we record too many things then it can become easy to lose track of the dependencies between them. The computation graph of a modern neural network can be deceptively large and complex, so it is important to organize all of your data in a user interface that is easy and intuitive to use.
We need the flexibility to look at the data from many different angles without having to spend too much time on writing code for the inspection, or worse, rerunning the model

\textbf{Comgra}.
Comgra \footnote{\label{comgra}Comgra: \url{github.com/FlorianDietz/comgra}
} is a library that aims to address these remaining problems.
It helps you inspect and analyze your network parameters as well as any tensors that are generated by your model, either as an output or as an intermediate step produced by a hidden layer.
It includes a GUI that makes it easy to understand the dependencies between different tensors and allows you to quickly switch between multiple different ways of looking at the data.
We took care to ensure that the amount of recorded data is kept to a reasonable level.
If you are interested in the average statistics over all data, but also in exact details for a specific input sample, Comgra can provide both in the same interface, all without a noticeable loss in performance.

\textbf{Usage}.
The library works similarly to Tensorboard: You record network activation tensors while the model trains. You then use a terminal command to open a browser window where you can inspect the results in a GUI.
This paper focuses on examples and usecases of comgra.
Please, see the website\footref{comgra} for installation and usage instructions, as well as code examples.

\subsection{Graphical User Interface}

\begin{figure*}[!htb]
\centering
\includegraphics[width=0.95\linewidth]{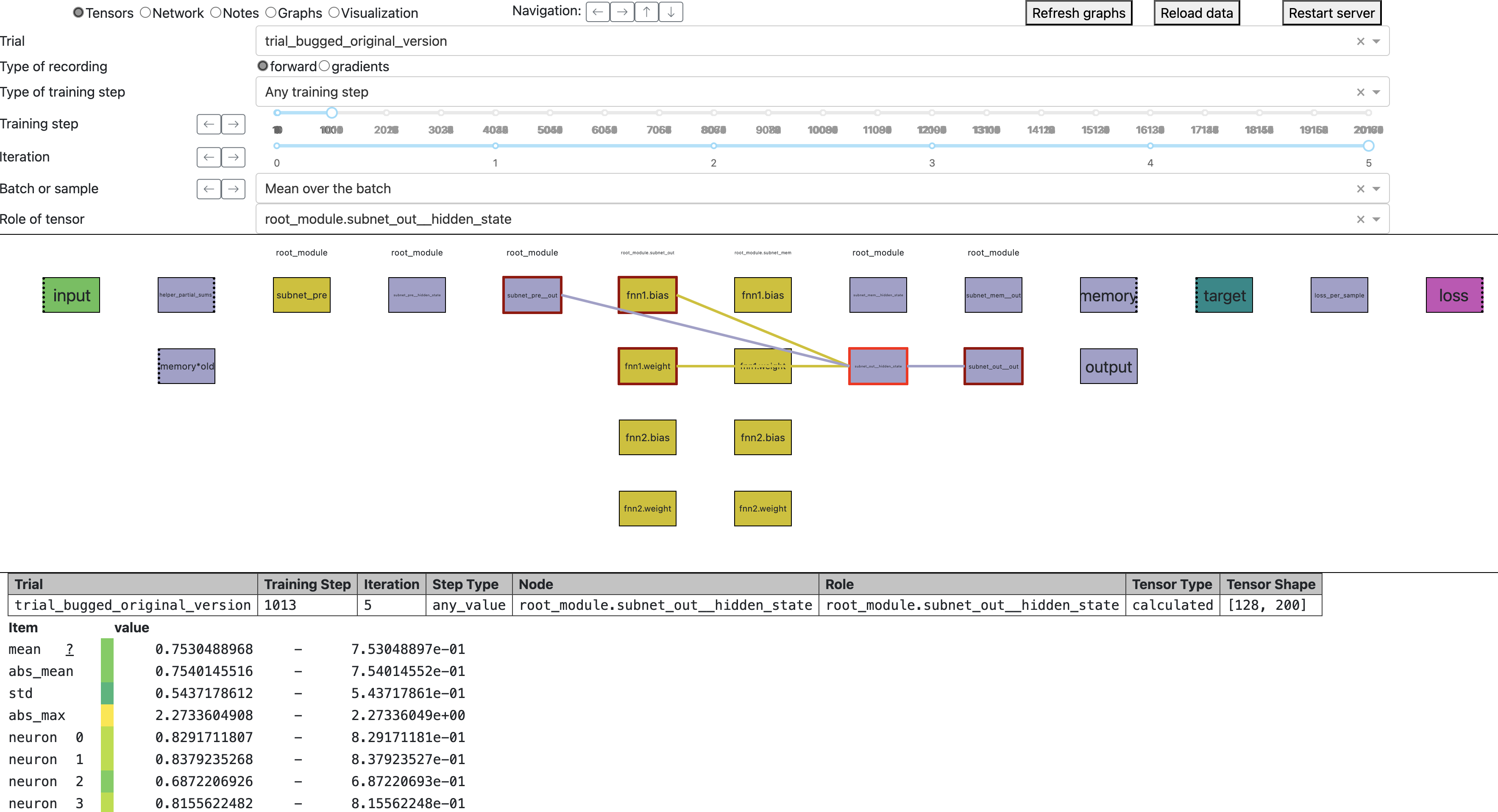}
\caption{Comgra's GUI. \textbf{Top}: Selectors for the version of the selected tensor and aspects you want to inspect. \textbf{Middle}: The dependency graph, in which you can click on a tensor to select it. \textbf{Bottom}: Summary statistics as well as raw values for the selected tensor under the selected criteria.}
\label{fig:gui-main}
\end{figure*}

Figure~\ref{fig:gui-main} shows the GUI. It consists of three parts: The selectors, the dependency graph, and the metrics.

\textbf{Selectors}
Comgra generates a number of versions of each named tensor over the course of a run and provides selectors to quickly switch between them.
\begin{itemize}
    \item Compare different trials, which may use different architectural variants.
    \item Select the training step.
    \item Filter the training step by a condition, e.g. only a step in which the target had a particular value, or only steps in which the input contained a particular rare token.
    \item Inspect either the tensor itself or the gradient on it. If you use multiple loss functions, you can also compare the gradients generated by each of them separately.
    \item Choose between viewing summary statistics over the batch or inspecting individual samples in the batch.
    \item Choose between variants if your model generates multiple instances of tensors with the same role in the dependency graph. For example, if you are using self-attention then comgra can store all tokens in one node and provides a selector to switch between the tokens.
\end{itemize}
Switching between these selectors is near-instantaneous even for large models with long training times. This allows you to rapidly test many different hypotheses without having to rerun your model.

\textbf{The Dependency Graph}.
Comgra automatically generates a dependency graph for all tensors it extracts. This graph is a subgraph of the computation graph that displays only the tensors you have chosen to log. This makes it easier to understand the graph because it lets you focus on the relevant parts.
It also makes it easier to compare different variants of architectures: Their computation graphs may be different, but the simplified dependency graphs are the same.
The dependency graph is automatically generated, but can also be customized to be more readable and easier to navigate if necessary.
Each rectangle in the dependency graph represents a named tensor that can be selected for inspection. The colors indicate the roles of the tensor in the network, such as input, parameter, calculated value, target and loss.
Note that the arrows between the nodes are only shown for the currently selected node. Due to the densely connected nature of computation graphs, this is easier to interpret in practice.

\textbf{Display}.
Comgra displays both the raw values of the selected tensor and summary metrics. The summary metrics are often all that is needed to detect outliers and anomalies. For a more in-depth analysis, it can be necessary to look at individual neurons or at how their values are distributed across a batch. You can use the Selectors to quickly switch between these types of information.

\textbf{Alternative Visualizations}.
The combination of selectors and dependency graph is the main benefit of comgra, but the GUI also includes some additional minor features because it is convenient to have them all available in a single tool. Use the radio buttons at the very top of the screen to switch to these: \textit{Network} shows the modules your network is composed of and gives a hierarchical breakdown of the number of parameters in them. \textit{Notes} displays simple textual log statements. \textit{Graphs} supports features similar to tensorboard\footref{tensorboard}, though it is simpler. Lastly, the \textit{Visualization} tab allows you to provide a custom visualization to comgra, through a python file using the \textbf{Dash}\footnote{Dash: \url{github.com/plotly/dash}} library. This visualization can depend on all the same filters and selectors that the rest of comgra uses, allowing you to apply comgra's flexibility with the specific visualization requirements of your task. For example, you might supply a script that can color-code the text fed into a transformer based on its attention weights and use comgra's selectors to investigate differences between different stages of training on the same samples.

\subsection{Dynamic Logging}

\textbf{Frequency of Recording}.
Comgra can dynamically adjust when to record a training step. You can record frequently at the beginning of training and less frequently as training goes on.
This allows you to get detailed results you can investigate early on during training without overwhelming your computer's memory if you keep the training process running for days.

\textbf{Categorizing Recordings}.
You can assign a \textit{Type of Training Step} to each input and make the decision to log for each type separately.
This means that a rarely encountered input will still appear in the logs frequently enough, even if the overall logging frequency is low. In this way you can ensure that the logs always contain representative samples for each type of input you care about.

\section{Usecases}
\label{section:example_features}

Comgra makes it possible to look at your data from many different points of view in a short period of time. \emph{The main advantage of the tool is its versatility and speed}.

In this section, we give a high-level overview of different features and usecases. If any of these pique your interest and you decide to give comgra a try, we recommend checking out the tutorial on the comgra website\footref{comgra}, which is too long to be reproduced in this paper. It illustrates how comgra can be used for debugging on an example task. The example is based on a real bug the authors found in one of their networks while developing a new architecture. In particular, it was a bug that led to a reduction in training speed but still allowed the network to learn. It is very unlikely that we would have noticed this without comgra.

The following is a list of different features of comgra. Some of these help with debugging, some with architecture optimization, and some with interpretability. Many of them help with multiple of these aspects at once.

\textbf{Getting an Overview}.
The first and often most useful thing you can do with comgra is to spend just a few minutes exploring your network to get an overview.
Look at the inputs, outputs, and targets for different batches and at different training steps. Is the target what it should be? Does the output approximate it well? Do intermediate tensors all have the same value range, or are some of them larger or smaller than others? Can you notice any irregularities when you compare different items within a batch?
When you look at the hierarchical breakdown of network parameters, do any of the modules have fewer parameters than others, forming a bottleneck?
Comgra's GUI makes it much easier and faster to perform the sanity checks than using conventional debugging tools, which can easily save you hours or days of frustration by catching a simple mistake early.

\textbf{Initialization}.
What do tensors and their gradients look like in the very first training steps? Do their means change significantly as training progresses? If they do, this suggests that they could be better initialized.

\textbf{Testing Toy Examples Step-by-Step}.
You can use the \textit{Type of Training Step} selector to record specific data points separately.
Use this to record toy examples and inspect all tensor activations of the network to see if they are in line with your expectations.
Comgra makes it much easier to get all the relevant details, which makes this much less time-consuming and therefore more practical.
In addition to helping with debugging, you can also use this for interpretability research:
By comparing the activations of specific examples of interest with the averages over other examples, you can find out very quickly if there are any neurons or statistics in the network that consistently have different values for a specific type of data point than for other data.

\textbf{Compare Categories of Data}
Similarly, you can use the \textit{Type of Training Step} selector to group your data by similarity and then use comgra to look for correlations.
You don't need to think of all possible hypotheses in advance: Simply create a set of recordings that seems like a reasonable breakdown of your data that you might want to investigate later.
You can rely on comgra to record a large enough variety of internal activations and statistics that you will be able to answer any questions that might arise later.
For example, when testing a transformer, you might create separate \textit{Type of Training Step} values for batches with particularly short inputs, or particularly long inputs, or for batches with particularly many commas relative to the sentence length.
Later on you may come up with a specific hypothesis such as ``I expect longer inputs to make more use of residual connections than shorter ones, and I expect this to be more pronounced later in training."
Even for such a specific hypothesis, comgra will already have extracted all the data necessary to answer that question. It takes just a couple of minutes of inspecting the right nodes in the dependency graph and adjusting the right sliders and selectors.

\textbf{Tracing the Origins of NaNs and Extreme Values}.
If NaNs or extreme outlier values occur anywhere in the network, you can simply follow the dependency graph to trace where they come from.
This is trivial for NaNs even without comgra, but with comgra it becomes possible to backtrack the source of unusually large but still valid numbers as well:
Use the selectors to find a particular element of the batch that is an outlier, then backtrack through the graph, and check at each node if its values at that node are outliers relative to the mean and standard deviation of the batch.
In this way, you can find the earliest part of the network where the data starts being unusual, even if it does not cause any numerical instabilities yet.

\textbf{Organizing Tensors}.
In complex neural networks, it can be easy to lose track of dependencies. Comgra helps with this by reducing the full computational graph to the dependency graph, but it also offers an additional feature to further simplify things.
A single node in the graph can store multiple tensors that fulfill different roles, and you can switch between them using the \textit{Role of Tensor} selector.
In this way you can, e.g. make every element in a variable-length Attention mechanism selectable without the visual clutter of creating one node per token. You simply switch between tokens using the \textit{Role of Tensor} selector, instead.
Furthermore, you can use this to display helper variables and derived values of the network without visual clutter. In our example script in the tutorial, we defined a node called ``helper\_partial\_sums", which can be found to the top left of the dependency graph. Unlike other nodes, this node contains several different tensors with different roles, and you can switch between them using the \textit{Role of Tensor} selector. We use this node to store the partial sums calculated during the example task, which would not normally get stored by the network but are helpful for debugging. We use the \textit{Role of Tensor} selector to switch between the different partial sums, so that all share a single node in the GUI, reducing visual clutter and making the data easier to find.

\textbf{Investigating Phase Shifts}.
Neural networks can exhibit strong differences between different stages of training.
For example, loss curves will sometimes stay largely unchanged for a long time before suddenly dropping sharply.
Grokking \citep{Power2022GrokkingGB} is an even stranger case, where test performance improves even though training performance remains unchanged.
Comgra allows you to investigate this. You can quickly compare how the values and statistics of each tensor change over time as training proceeds. Are the values for two training steps within the same phase different from the training steps in another phase?
Even without a concrete hypothesis in mind, you may notice commonalities that then serve as the inspiration for a deeper analysis.

\textbf{Architecture Design}.
You can easily compare different variants of an architecture using the \textit{Type of Training Step} selector.
If the differences are minor, you may even find that the dependency graph is the same, even though the computation graphs are different. This makes a one-to-one comparison possible.
If the differences are localized, you can inspect the statistics at the nodes before and after that location in the dependency graph.

\textbf{Variance over the Batch: Mode Collapse and Infinite Growth}.
We have found that inspecting the variance over the batch of any tensor is a very simple way to detect common problems.
If the variance decreases, we may have mode collapse.
If it keeps increasing over time, it suggests that a tensor is receiving consistently one-directional gradients that make the tensor more and more extreme instead of coming to approximate a target value more accurately.
The latter problem is less catastrophic, making it all the more important to detect: If your network suffers from mode collapse, then you will notice this as performance drops. But a steadily growing value means that the network can still solve the problem, but is less efficient than it could be, which is very difficult to detect.

\textbf{Interventions}.
If I perform an ablation, for example, by setting an intermediate activation to zero, what effects does this have on other network weights?
It can be difficult to predict which parts of the network will be affected by such an intervention.
You can use the \textit{Type of Training Step} selector to indicate interventions on the model. You can then inspect tensors and the statistics on them while switching between the intervention and the normal run.
This lets you quickly find out how different parts of the network are affected by an intervention.

\textbf{Exploding and Vanishing Gradients}.
Comgra allows you to look at the gradients of your model: The GUI has a selector to switch from forward mode to gradients.
If you notice exploding or vanishing gradients, you can find out which calculation is causing it by simply clicking through the node of the dependency graph until you have found the rightmost tensor with anomalous gradients.

\textbf{Imbalanced Gradients}
You can also use comgra to detect more subtle issues with gradients:
All the same statistics are available for the gradients that are also available for the tensors themselves. This makes it easy to detect when e.g. the training data has rare outliers with abnormally high gradients, because those will result in a high variance of the gradient over the batch.
You can also use the dependency graph to compare how much the gradients from different computation paths contribute to a shared ancestor.
If a tensor $A$ is a dependency of tensors $B$ and $C$, and the average absolute gradient on $B$ is a hundred times as large as on $C$, then this imbalance suggests room for improvement.
$C$ will still learn under these conditions, but much more slowly than it could, because $A$ will be biased to develop its weights mostly to support $B$ and not $C$.
This is the kind of issue that slows down your model but does not break it, which is in general very difficult to discover.

\textbf{Finding Interpretable Neurons}.
You can use comgra to check if any neurons end up with interpretable values. For example, the weights in attention mechanisms tell you what the network pays attention to. But there are also more subtle interpretable values that would be difficult to inspect without comgra, unless you already know what to look for before you run the experiment. For example, you can compare the mean absolute values of the two branches in a residual connection to find out if the network ignores a calculation and relies on residuals.
In most cases, neurons will not be interpretable. But you will not know it until you try, and comgra's GUI makes it easy to inspect your data from a lot of different angles in a short amount of time.

\section{Future Work}

\textbf{Improved Dynamic Logging}.
Comgra's dynamic logging has some limitations. You currently need to make the decision whether to log and what type to assign to the log before the training step starts. This means that you can create separate logs for particular inputs that you have selected ahead of time, but you cannot change the decision to log based on the results of the batch. We aim to enable the decision to log only after the results of the batch are seen, and to focus on specific samples of that batch based on their properties. This will allow you to detect and focus more easily on particular cases.
For example, while training a language model, you might want to review only the adversarial samples that pose a security issue to the model.
More generally, if you combine this feature with anomaly detection on standard KPIs, then comgra will be able to extract outlier samples, making them easier to investigate.
Note that this feature is being implemented at the time of this writing and will likely be finished by the time you are reading this.

\textbf{Anomaly Detection}.
A goal for the future development of this tool is the automated detection of anomalies in computation graphs.
It should be possible to define anomalies like ``Tensor X has a greater absolute value than 1" or the like, and then have the program automatically calculate likely dependencies such as this:
The anomaly ``abnormally high loss" has 87\% correlation with the anomaly ``Tensor Y is close to zero".
This would save a lot of time with debugging by automatically generating a list of possible reasons for unexpected behavior.
Similarly, it could be used for interpretability research by automatically scanning the network for correlations between the categories of data points used and for anomalies in the network activations.
The goal of this feature is not to detect anomalies with perfect reliability, but to quickly and cheaply generate hints that guide a human's attention in the right direction, to save time.


\section{Conclusion}

Comgra provides many different features to help you explore the internals of your neural network.
It logs a diverse set of data while maintaining a low memory footprint and computational overhead.
It enables the user to inspect their network in a GUI from many different angles while remaining easy to use.
This makes it useful both for explorative analysis and for quickly testing new hypotheses about network behavior without having to rerun the network.
We have shown usecases that demonstrate its usefulness for debugging, architecture optimization, and interpretability.

\section{Author Contributions}

Comgra was invented and developed by Florian Dietz. Professors Sophie Fellenz, Marius Kloft and Dietrich Klakow provided valuable feedback on the paper and helped acquire users for testing the library.

\section{Acknowledgements}

Work by the author was supported by a grant by the NHR-Verein (National High Performance Computing, Germany).

Comgra relies on PyTorch for training neural networks. It uses Dash\footnote{Dash: \url{https://github.com/plotly/dash}} as the basis of its GUI. We would like to thank their creators for these valuable tools.

\bibliography{main}
\bibliographystyle{icml2024}

\end{document}